\documentclass[
    final
    ,5p
    ,times
    ,twocolumn
    ,authoryear
]{elsarticle}

\usepackage{algpseudocodex}
\usepackage{mdframed}
\usepackage{newfloat}
\usepackage{url}
\usepackage{csquotes}
\usepackage[hidelinks]{hyperref}
\usepackage{pbalance}

\DeclareFloatingEnvironment{example}
\journal{Cognitive Systems Research}

\begin{document}

\begin{frontmatter}

\title{%
    Forms of Understanding for XAI-Explanations
}

\author[ubi,trr]{Hendrik Buschmeier\corref{ca1}}
    \cortext[ca1]{Corresponding author.}
    \ead{hbuschme@uni-bielefeld.de}
\author[upb,trr]{Heike M. Buhl}
\author[ubi,trr]{Friederike Kern}
\author[upb,trr]{Angela Grimminger}
\author[ubi,trr]{Helen Beierling}
\author[upb,trr]{Josephine Fisher}
\author[ubi,trr]{André Groß}
\author[uwk,trr]{Ilona Horwath}
\author[upb,trr]{Nils Klowait}
\author[upb,trr]{Stefan Lazarov}
\author[upb,trr]{Michael Lenke}
\author[ubi,trr]{Vivien Lohmer}
\author[upb,trr]{Katharina Rohlfing}
\author[upb,trr]{Ingrid Scharlau}
\author[upb,trr]{Amit Singh}
\author[upb,trr]{Lutz Terfloth}
\author[ubi,trr]{Anna-Lisa Vollmer}
\author[ubi,trr]{Yu Wang}
\author[upb,trr]{Annedore Wilmes}
\author[ubi,trr]{Britta Wrede}

\affiliation[ubi]{organization={%
    Bielefeld University},
    city={Bielefeld},
    country={Germany}}

\affiliation[upb]{organization={%
    Paderborn University},
    city={Paderborn},
    country={Germany}}

\affiliation[uwk]{organization={%
    University for Continuing Education Krems},
    city={Krems},
    country={Austria}}

\affiliation[trr]{%
    organization={SFB/Transregio 318 ‘Constructing Explainability’},
    city={Paderborn \& Bielefeld},
    country={Germany}}

\begin{abstract}
    Explainability has become an important topic in computer science and artificial intelligence, leading to a subfield called Explainable Artificial Intelligence (XAI). The goal of providing or seeking explanations is to achieve (better) ‘understanding’ on the part of the explainee. However, what it means to ‘understand’ is still not clearly defined, and the concept itself is rarely the subject of scientific investigation. This conceptual article aims to present a model of forms of understanding for XAI-explanations and beyond. From an interdisciplinary perspective bringing together computer science, linguistics, sociology, philosophy and psychology, a definition of understanding and its forms, assessment, and dynamics during the process of giving everyday explanations are explored. Two types of understanding are considered as possible outcomes of explanations, namely \emph{enabledness}, ‘knowing how’ to do or decide something, and \emph{comprehension}, ‘knowing that’ -- both in different degrees (from shallow to deep). Explanations regularly start with shallow understanding in a specific domain and can lead to deep comprehension and enabledness of the explanandum, which we see as a prerequisite for human users to gain \emph{agency}. In this process, the increase of comprehension and enabledness are highly interdependent. Against the background of this systematization, special challenges of understanding in XAI are discussed.
\end{abstract}


\end{frontmatter}

\section{Introduction}
\label{sec:introduction}

Due to recent technological developments, explainability has become a widely discussed topic in computer science and artificial intelligence, giving rise to a new subfield called Explainable Artificial Intelligence (or XAI for short). It is also a topic of interest in other disciplines, such as cognitive science, psychology, (psycho-)linguistics, philosophy and education. The focus of this article is to provide a theoretical concept of understanding as a result of explanation. From this perspective, understanding is the goal of a successful explanation \citep[p.~229]{Keil2006} and \enquote{recipients of explanations, if the explanations are at all successful, are expanding their understanding in real time}. Understanding has implicit knowledge parts \citep[p.~97]{WilsonKeil2000} regarding its procedural and non-reflective components. As explanations progress, the level of understanding increases, but understanding always remains incomplete \citep{Miyake1986} and is not always factual \citep{Paez2019}, among other reasons because explanations are often shallow, depending on the knowledge and motivation of the explainer and the explainee \citep{WilsonKeil2000}, and they have \enquote{relatively unclear end states} \citep[p.~242]{Keil2006}. 
Thus, while there is thus a broad agreement that the goal of explanation is understanding, we know little about how understanding can be conceptualized, what forms of understanding can occur, and, finally, how it can be measured. The aim of this article is to bring together insights from several disciplines and to develop a concept based on their findings. We will propose a \emph{four-field model of understanding} that provides conceptual clarification of related terms and concepts, and thus distinguishes between different forms and levels of understanding. Our goal is to develop a model that can be used as a heuristic to capture, analyze, and even measure different forms of understanding in the context of XAI and beyond, and that can be operationalized in interactive explanation systems.

The starting point of our considerations is a view of an XAI-explanation as a co-constructive, interactive process to which both humans and XAI or interactive explanation systems contribute \citep{RohlfingCimiano2021, Rohlfing2025socialxai}. The goal of an explanation -- i.e., the form and/or level of understanding it aims to achieve -- depends on the goal of the explainer as well as contextual and situational contingencies \citep{RohlfingCimiano2021}. Rather than focusing solely on the narrow case of \emph{explainability}, i.e., understanding the inner workings of an AI system using XAI methods, we consider XAI to have a broader scope of \emph{explaining} more generally (e.g., explaining a programming language concept to a novice computer programmer, a grammar rule to a second language learner, or a board game to someone learning to play it) fostering understanding and agency of humans.
Accordingly, we adopt a broad notion of ‘understanding’ as implemented in and underlying everyday interactions and experiences with algorithmic technology in socio-technical AI systems. In contrast to other research on explanation, we do not focus on scientific explanations but on ‘everyday explanations’ because they show a greater variety with regard to the knowledge they mediate \citep{Keil2006}. While scientific explanations are regularly characterized by \emph{why}-questions \citep[e.g.,][]{ChideLeeuw1994}, and thus aim primarily to convey causal knowledge, everyday explanations often follow other kinds of questions, such as \emph{what}, \emph{where}, \emph{how does it work}, etc. \citep{Klein2009}. Moreover, the quality of everyday explanations is more difficult to define -- they can be very simple, incomplete, inaccurate, or, in contrast, quite sophisticated \citep{Johnson-Laird1983, Norman1988}.

The above-mentioned disciplines have dealt with the problem of understanding to varying degrees. While understanding is a widely discussed topic in philosophy, it is not a topic in experimental psychology nor is there a scientific concept of it. Likewise, ‘understanding’ is hardly considered a research topic in its own right in many disciplines \citep[but see, e.g,][]{Zagzebski2019}. Nevertheless, this article will provide a synthesis of some interdisciplinary perspectives on ‘understanding’ as a result of everyday explanations, i.e., non-scientific explanations that occur in everyday contexts and can involve either humans or a human and a machine. The objective is to present a model of different \emph{forms} of understanding that can arise from them. This is highly relevant for tutoring contexts \citep[e.g.,][]{ChiSiler2004, Miyake1986}, classroom interaction \citep[e.g.,][]{Koole2010, vandePolMercer2018, MartinsGresse2022}, as well as for XAI \citep{RohlfingCimiano2021}. We would like to argue that a model that attempts to define and unify basic concepts is a useful tool across disciplines because it can provide a starting point for both further empirical analysis and theory development.
While the proposed model is primarily concerned with cognitive aspects, we will also argue that embodied interaction should be included. With such a claim, we intend to expand the notion of understanding from a purely cognitive conception to one that includes bodily aspects. The model will be enriched by offering a theoretical account of the assumed dynamics of understanding by sketching how movement between the four fields can be conceptualized.

\subsection{Basic concepts and first examples}
\label{sec:basic-concepts}

In this paper, \emph{explaining} is conceptualized as a social interactive process \citep{Miller2019, Miller2023} in which an \emph{explainer} and an \emph{explainee} iteratively \emph{co-construct} an \emph{explanans} of an entity that is explained (the \emph{explanandum}). This concept is based on everyday explanations which, in contrast to scientific explanations, are always interactive and usually embedded in a situational context. The explainer is an agent that can either be a human or, as it is the case in XAI-explanations, an interactive explanation system. Both partners are actively involved in the explanatory process through mutual monitoring and scaffolding \citep{RohlfingCimiano2021} and proposals have been made how to model such activities for interactive explanation systems \citep{RobrechtKopp2023, BooshehriBuschmeier2024, BuschmeierKopp2018}. The goal of an explanatory process is for the explainees to \emph{understand} the explanandum with regard to their situational objectives or purposes; the form and depth of understanding, however, can vary according to the explainee's goals as well as situational and contextual contingencies. 
This concept of explaining that emphasizes the process rather than the (normatively fixed) outcome differs from the view often held in science that explanations are products with formalized normative forms, and comply to rules of logic \citep{Grimm2021-SEP}. Everyday explanations also include causal, procedural or conceptual relations \citep{QuasthoffHeller2017}, but usually do not follow forms or rules of natural science; neither do they necessarily aim at formulating universal (scientific) laws. Instead, everyday explanations are more tailored to the explainee's situational objectives and needs. They are therefore more context-dependent than scientific explanations and correspondingly more flexible in terms of form and content \citep[cf.,][]{Miller2019}.  

In this article, we will focus on ‘understanding’ as knowledge resulting from a given explanation. This should not mean that explanations always lead to understanding; indeed, they might lead to a false impression of understanding, or they may lead to no understanding at all, regardless of the explanation's quality \citep{Hempel1965}. In fact, the connection between explanations and understanding is still of debate in philosophy. While some argue that understanding is conceptually prior to explanations, others claim that understanding is in fact knowing what a good explanation is, and the differences between understanding and explanatory knowledge are only minor \citep{Khalifa2017}. While this is an interesting debate in itself, our focus here is different, namely to present a heuristics for forms of understanding which can arise as a result of successful knowledge transfer through everyday as well as XAI-explanations. We will thus not go into the relation between the two any further; instead, we assume for theoretical convenience that explanations can lead to knowledge transfer and thus to understanding on the part of the explainee.

Before presenting a conceptual model of understanding, we assume that the notion of knowledge should be clarified. As mentioned above, we follow the view that it is an explanations' objective to construct and transmit knowledge \citep{Miller2019, QuasthoffHeller2017}, and we assume that understanding something on the basis of an explanation means knowing something. It should be noted that there is a debate in philosophy about differences between ‘understanding’ and ‘knowledge’. Knowledge, it is argued, can be rather easy to acquire, and objects of knowledge can be isolated \citep{Grimm2021-SEP}. Understanding, in contrast, seems to be more difficult to get because it involves connecting isolated knowledge objects with each other and combining them into a structured whole \citep{Zagzebski2019}. For the purpose of this article, we neglect the debate about possible differences between understanding and knowledge and take the view that the differences are marginal at best.
However, building on previous scholarly work on knowledge and learning, a social constructivist view underlines the importance of the subjective parts of knowledge that arise from different goals, prior knowledge, etc. \citep[cf.,][]{palincsar1998social}. If two people read the same book, they will not have the same knowledge afterwards. Consider, for example, the geocentric worldview with the earth at the center of the universe, which was socially shared ‘knowledge’. Knowledge therefore also has subjective components. This is especially the case when we talk about the knowledge that results from understanding everyday explanations.

The different forms of understanding that an explainer is aiming for -- or that are sought after by the explainee -- are reflected in the heterogeneity that can be found especially in everyday explanations. In order to deal with this diversity, we have developed a model that considers two types of understanding as the result or goal of an explanation: \emph{comprehension} and \emph{enabledness}. In addition, the model conceives of these types as further differentiated by the level of understanding that can be achieved, which we label \emph{shallow} and \emph{deep} respectively \citep[cf.][p.~127]{GraesserPerson1994}. It should be noted that our model, like any model, is simplified \citep{Bailer-Jones2009} in order to provide a systematic examination of different forms. 

The term \emph{comprehension} can be roughly translated as ‘knowing that’. This refers to semantic or even theoretical knowledge that can be verbalized (‘declarative’ knowledge). At best it is a conceptual framework for a phenomenon that even goes beyond what is immediately perceivable. Regarding the distinction between shallow and deep levels of understanding, we assume that forms of understanding can also be either very simple/shallow forms, or rather deep. 

In contrast, the term \emph{enabledness} means ‘knowing how’. Again, enabledness can be very simple (‘to start an Internet search, enter the search term in the browser’) or deep, e.g., knowing how to program a computer game. Thus, a person can be enabled to perform either a simple routine action or a complex action. However, complex actions regularly require some form of comprehension to be enabled. We suggest that comprehension and enabledness are generally interdependent. 

In this article, we will use two examples to illustrate understanding (see examples~\ref{ex:while} and \ref{ex:aspect}): In the following, we would like to explain the terms introduced above using example~\ref{ex:while}. It looks at understanding in a technical context, namely the use of a stopping criterion in an iteration loop of a computer program. More specifically, example~\ref{ex:while} is an illustration to learn how to use \emph{while-loops} in computer programming. While-loops are constructs of programming languages that execute the code within the construct as long as the while-condition attached to the construct holds. In the example, a variable \texttt{counter} is decremented by $1$ as long as its current value is greater than $1$. The specific application of the example is to print the well known ‘99 bottles’ song\footnote{\url{https://en.wikipedia.org/wiki/99_Bottles_of_Beer}} \citep{Byrd2010}. While-loops are known to cause problems for computer programming students \citep{QianLehman2017} and could be a topic for an interactive explanation system in the domain of computer science education.

\begin{example}
\centering
\small
\begin{mdframed}
\begin{algorithmic}[1]
\State $counter \gets 99$
\While{$counter > 0$} 
    \State \Output $counter$ “ bottles of liquid on the wall,”
    \State \Output $counter$ “ bottles of liquid;”
    \State \Output “If one of those bottles should happen to fall,”
    \State $counter \gets counter - 1$
    \State \Output $counter$ “ bottles of liquid on the wall.”
\EndWhile
\end{algorithmic}
\end{mdframed}
\vspace{-3mm}
\caption{A simple example illustrating a while-loop.}
\label{ex:while}
\end{example}

\begin{description}

    \item[Shallow Comprehension] of the program would entail that it will print out the ‘99 bottles’-song on the computer screen. Specifically, regarding the while-construct, shallow comprehension could involve understanding that the indented lines (3–7) are executed every time the while-loop is entered and that \texttt{counter $>$ 0} in line 2 is the condition under which the while-loop is executed.

    \item[Deep comprehension] would go beyond this and mean to understand that the condition of the while-loop actively interferes with the while-loop: the variable \texttt{counter} is decreased and the condition reacts to this. Deep comprehension may also detect a shortcoming of the program, i.e., that the lines of the song are not printed correctly when the counter equals one, as the grammatical agreement (number) between number and noun is violated (*“1 bottles”). Going beyond this program, deep comprehension would also give the explainee the insight that the while-condition could also be used for different applications, e.g., to try to access an internet resource over an unreliable connection until the full resource has been downloaded.

    \item[Shallow enabledness] would allow an explainee to change the program, so that it would, for instance, start at 77 bottles or decrease by 2 each step or stop early. 

    \item[Deep enabledness] would allow an explainee to resolve the ‘1 bottles’-problem mentioned above by rearranging lines and combining multiple while-loops.

\end{description}

Example~\ref{ex:aspect} examines understanding in a non-technical, natural context, namely the use of the \emph{present progressive tense in English} by native and non-native speakers. While native speakers (at least of certain variants or English) may use and understand this construction intuitively, it can be challenging for learners of English as a second language. For these learners, the concept may need to be explained in a classroom setting or by an interactive explanation system in the context of language teaching. Understanding of such explanations might also be analyzed with respect to comprehension and enabledness.

\begin{example}
\centering
\begin{mdframed}
    This example looks at a specific construction of natural language use, specifically the English \emph{present progressive} tense. The present progressive is constructed using \enquote{the present tense form of be + lexical verb in -ing form} \citep[p.~598]{CarterMcCarthy2006} and is used for, among other things, “events in progress at the time of speaking”, “repeated events in temporary contexts”, “processes of change”, and “with adverbs of indefinite frequency” \citep[p.~601f]{CarterMcCarthy2006}, i.e., it expresses aspect grammatically. These usage contexts may be intuitive to native English speakers, who may also be able to see the nuanced differences in usage compared to the \emph{present simple} tense. However, learners of English as a second language, especially speakers of languages where there is no grammatical aspect (e.g., standard German), need to understand how and when to use the present progressive and may not even have the concepts (e.g., that progressivity is grammatically marked) to begin with.
\end{mdframed}
\vspace{-3mm}
\caption{Grammatical aspect in natural language use.}
\label{ex:aspect}
\end{example}

\begin{description}

    \item[Shallow comprehension] of the present progressive would, for example, entail that the explainee, a learner of English as a second language, is aware that the present progressive is a tense that expresses events in progress and has the form am/are/is and a verb + -ing.

    \item[Deep comprehension] could mean that an explainee also understands the reasons why the present progressive should (not) be used in certain contexts and that it is use to express a certain grammatical aspect, namely progress.

    \item[Shallow enabledness] of the present progressive would allow a learner to correctly put sentences in the progressive tense without necessarily knowing when it is appropriate or not.

    \item[Deep enabledness] allows an explainee to use the present progressive competently when formulating utterances or sentences in speech and writing.

\end{description}
Section~\ref{sec:forms-of-understanding} will look at both examples in more detail, Section~\ref{sec:classic-xai-example} will apply the model to a classic XAI example.

\section{Aspects of Understanding: Forms of knowledge}
\label{sec:aspect-of-understanding}

For our model of understanding, we follow primarily two conceptualizations: the types and qualities of knowledge as described in detail by \citet{deJongFerguson-Hessler1996}, and the taxonomies of learning from the field of education \citep[‘Bloom's taxonomy’;][]{AndersonKrathwohl2001, KrathwohlBloom1969}. 

\Citet{deJongFerguson-Hessler1996} were interested in knowledge in instruction and learning and especially as a prerequisite for problem solving. They distinguished and elaborated characteristics of different types of knowledge. In terms of understanding as a goal and the achievement of explanations, two types of knowledge are important: \emph{conceptual knowledge} is \enquote{knowledge of facts, concepts, and principles} (p.~107), whereas \emph{procedural knowledge} refers to \enquote{actions and manipulations} that transform one state into another. Thus, in our terminology, conceptual knowledge describes ‘knowing-that’, and procedural knowledge refers to ‘knowing-how’. In the context of XAI, we focus on cognitive forms of procedural knowledge, such as written division and programming. Actions and skills with a higher motor component, such as cycling or knitting, may require other forms of more embodied explanations (showing). 

Despite the functional proximity between the notions of conceptual knowledge and comprehension, and procedural knowledge and enabledness, we prefer to use both \emph{comprehension} and \emph{enabledness} in the context of everyday explanation to emphasize their association with explanation. The achievement of explanations often goes beyond a specific domain and situation because explanations are tailored to the goals of the explainee and the situation. With the term ‘comprehension’ we also relate to the work of \citeauthor{Kintsch1998} (e.g., \citeyear{Kintsch1998}), who uses ‘comprehension’ for the process of comprehension in the area of text processing.

Building on the taxonomy of \emph{qualities of knowledge} also developed by \citet{deJongFerguson-Hessler1996}, we consider the following characteristics to be relevant for describing understanding:

\begin{description}

    \item[Level] Both conceptual and procedural knowledge are described as more or less shallow or deep. Regarding conceptual knowledge, \citet[p.~111]{deJongFerguson-Hessler1996} illustrate the distinction with symbols or formulas on a shallow level vs. concepts and their relations on a deep level. Thus, deep comprehension most likely reflects highly interconnected and elaborated concepts. Shallow comprehension would correspond to little interconnectedness, but also to few concepts. In terms of procedural knowledge, there are rules or recipes at a shallow level in contrast to meaningful actions at a deep level. 

    Explainees may require different levels of understanding, and explanations should be tailored to those needs \citep{WilsonKeil2000}. Thus, it is not always necessary -- or even annoying -- to give or receive a long and complex explanation that leads to deep understanding. Often the explainer is satisfied with a single action or a simple fact.

    \item[Structure] Knowledge can be more or less structured, i.e., singular elements of knowledge can be isolated or represented in a meaningful structure. This structure facilitates the retrieval of knowledge and the acquisition of new knowledge. Semantic networks, for example, are used to visualize knowledge structures. The distinction between knowledge and understanding of novices vs. experts (see Section~\ref{sec:novice-expert}) can be captured by describing how elements might be structured: The knowledge structure of experts is more hierarchical and better organized than that of novices. Although level and structure are closely related, an explanation can also lead to a false structure at a very deep level.

    \item[Automation] Automation of knowledge is particularly important for understanding and therefore for explanation. Consider how difficult it is to explain simple skills like riding a bicycle. This is because there is a strong distinction between declarative conceptual knowledge and non-declarative compiled procedural knowledge. Compilation enables automatic processes. The term “knowledge compilation”, inspired by computer science, is used by \citet{Anderson1983} to describe how multiple cognitive steps are combined into a single routine. Compilation involves two stages: composition, where steps are combined, and proceduralization, where declarative knowledge becomes automatic, resulting in more fluid performance.
    
    We conceive comprehension as involving conceptual knowledge, i.e., remembering facts about and experiences with various concepts such as theories and/or related routine actions. In contrast, enabledness is conceived as the ability to perform routine and/or complex actions (mentally and physically) in a competent and flexible manner. Furthermore, whereas comprehension is always ‘explicit knowledge’ that can be acquired through books or explicit verbal instruction, enabledness consists at least in part of ‘tacit knowledge’ that is not easily shared \citep{LimAhmed2000}.

    However, conceptual knowledge can also be compiled, e.g., rote learning of definitions or mnemonic sentences instead of deep comprehension. On the other hand, deep enabledness requires declarative, mostly deep comprehension. In addition to implicit learning of procedures as a basis for skills, e.g., native language grammar, there is explicit learning, e.g., adult learning of a foreign language. Explaining how to drive a car is a good example of how procedures are first explained in a declarative way, and then the steps are proceduralized and composed over the years \citep{Anderson1983}. \Citet{deJongFerguson-Hessler1996} explain automation by mentioning tacit knowledge that cannot be expressed. In order to give an explanation, the explainer needs non-tacit knowledge, also of procedures. In the following sections, this intertwining of understanding quality and type will be discussed in detail.

    \item[Modality] Regarding modality, we added embodiment to the verbal and pictorial aspects given by \citet{deJongFerguson-Hessler1996}. Embodied knowledge is conceptually similar to procedural knowledge \citep[p.~150]{Tanaka2011}, and it can be better performed than explained verbally. Embodiment plays an important role in learning situations where new members of a (scientific) community are taught to become skilled practitioners. Visibly embodied displays (e.g., how to hold a pencil, how to hold a Munsell color chart next to a piece of dirt to describe its color) are part of the knowledge conveyed in such situations \citep{Goodwin2013}.

    \item[Generality] Regarding generality, explanation and understanding are domain specific. However, it is important to keep in mind for the following discussion that an explainee may be a novice with no understanding in one domain, but an expert in another. Moreover, the given domain can be very specific (e.g., the motivation of a particular person in a particular situation) or on a more general level (e.g., the motivation of people -- even living beings).

\end{description}

In addition to this characterization of different types of knowledge and their manifold qualities, as a further conceptualization, taxonomies of learning objectives help to describe the types and levels of understanding \citep[e.g.,][]{KrathwohlBloom1969,AndersonKrathwohl2001}. \citealt{AndersonKrathwohl2001} (summarized in \citealt{Anderson2002}) systematize educational objectives in two dimensions, the structure of the knowledge and the cognitive processes that are to be done with the knowledge.

\begin{description}

\item[Structure of knowledge]
    The structure of knowledge aims at a differentiation of the above mentioned ‘types of knowledge’. It includes factual, conceptual, procedural, and metacognitive knowledge. While factual knowledge contains the basic elements of a domain (e.g., definitions), conceptual knowledge contains the structure of the basic elements, e.g., principles, models, and theories. Comparable to \citet{deJongFerguson-Hessler1996}, procedural knowledge aims at ‘how to do something’, such as skills and algorithms. The fourth form of knowledge, metacognition, will be neglected in our paper, because it is only of interest in special explanations regarding (own) cognition. Thus, the categories of knowledge are similar to the types of knowledge discussed by \citet{deJongFerguson-Hessler1996} with the additional hint that there are very basic elements of knowledge. In terms of understanding, factual and conceptual knowledge aim at comprehension, while procedural knowledge, of course, aims at enabledness.

\item[Cognitive processes]
    The second dimension of the taxonomy of learning aims at increasing the levels of \emph{cognitive processes} that can be done with knowledge, namely remembering, understanding (note that this is a different use of the term), applying, analyzing, evaluating, and creating \citep{ChiWylie2014}. This strengthens the aspect of the level of knowledge introduced by \citet{deJongFerguson-Hessler1996}. There is a gain in complexity from low level cognitive processes like remembering to applying to high level processes like creating something new -- resulting in a hierarchy from shallow to deep understanding. However, assessing the level of understanding makes it necessary to take a closer look at these processes. For example, whether creating something new is a high-level process could depend on what is meant by ‘new’ and the complexity of what is being created, e.g., a new sentence or a scientific article. 

\end{description}

\begin{figure}
    \includegraphics[width=\columnwidth]{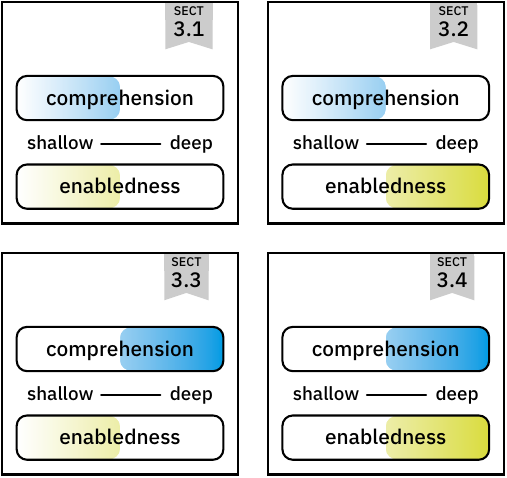}
    \caption{%
        Forms of understanding as a four-field diagram. Understanding can be of the type \emph{comprehension} and/or \emph{enabledness}, each of which has two levels: \emph{shallow} and \emph{deep}. The four fields are based on the possible combinations: e.g., shallow comprehension and shallow enabledness (top left, Section~\ref{sec:scse}), shallow comprehension and deep enabledness (top right, Section~\ref{sec:scde}), deep comprehension and shallow enabledness (bottom left, Section~\ref{sec:dcse}), and deep comprehension and deep enabledness (bottom left, Section~\ref{sec:dcde}).}
    \label{fig:types-and-levels-of-understanding}
\end{figure}

Building on these taxonomies, we focus on the two types of understanding introduced above (comprehension and enabledness) and the different qualities (levels) associated with them. We will use the terms shallow and deep comprehension on the one hand, and shallow and deep enabledness on the other. The conceptualization in the form of four fields -- as sketched in the diagram in Figure~\ref{fig:types-and-levels-of-understanding} -- is only for simplicity. We see the main advantages of the model in its ability to map the process of understanding at the different stages of the explanation process.

To avoid misunderstandings of the four-field diagram, it is important to note that the distinction between shallow and deep understanding is not dichotomous -- there are fluid transitions between the two extremes -- nor are comprehension and enabledness assumed to be independent (orthogonal) categories. The level of depth depends on the domain, the given explanation, and the goal being pursued. Accordingly, there is no static definition of deep understanding; instead, a change in depth is likely to occur as soon as either the goal or the domain is changed. These characteristics are discussed in the following.

In summary, there are two forms of understanding -- comprehension and enabledness. Both forms of understanding can be on different levels, from very shallow to very deep understanding. The four resulting combinations (cf. Figure~\ref{fig:types-and-levels-of-understanding}) are intermediate steps or final achievements of explanations with the possible goal of agency, the ability to control and adapt the (digital) environment (\citealt{PasseyShonfeld2018}; Section~\ref{sec:dcde}). The combinations will be characterized in the next section. On this basis, (in Section~\ref{sec:dynamics-of-understanding}) we explain dynamics of comprehension in explanations, taking into account two relevant aspects: We consider differences between novices and experts in explanations and discuss how in the course of an explanation comprehension and enabledness are intertwined.

\section{Forms of Understanding}
\label{sec:forms-of-understanding}

In the following, we will explain how understanding can be characterized for each of the four ‘fields’ on the basis of examples~\ref{ex:while} and \ref{ex:aspect} introduced above. The respective knowledge is clarified additionally by means of its assessment. Due to their subject matter and also the amount of available research on it, the sections differ in their extent.

\subsection{Shallow comprehension, shallow enabledness}
\label{sec:scse}

We see the combination of shallow comprehension and shallow enabledness as a potential starting point for understanding. Thus, this section will focus on the mental state before a change in understanding, or here, before explanation. The first step is to define the underlying conditions and premises of this stage. Any process of understanding should begin with and be tailored to the explainee's mental representation model, which is built from (prior) own and mediated experiences with the domain and prior knowledge about familiar things. 
At the beginning of an explanation, explainers and explainees need to establish 'common ground' \citep[p.~109]{Clark2014}, which involves assessing \enquote{what the other person knows at the start, then tailoring one's own contributions to add to the relevant common ground for that occasion […]}. Common ground is usually based on the prior knowledge that explainees bring to the explanation process. Prior knowledge thus influences the course of the explanation and the understanding that follows: Explainers can build their explanation on it, and explainees' mental processing of new knowledge is influenced by it \citep{ChideLeeuw1994}. Therefore, it is particularly interesting to know what prior knowledge is and which parts of it are relevant for an explanation, and how the explainer can identify these relevant parts of prior knowledge. To clarify the situation in this field, we begin with the two examples.

\paragraph{Example~\ref{ex:while}}
An example of people having shallow comprehension and shallow enabledness of a stopping criterion in an algorithm might be novice programmers. Their level of understanding might, for example, be characterized by prior knowledge of and basic familiarity with the syntax of the conditional statement in while-loops and using it to control for a specific number of iterations (e.g., do this exactly three times), but not knowing (yet) that it can also be used in a more dynamic way (\texttt{a < b}) (shallow comprehension) or how to set up such conditions generically (shallow enabledness). Of course, novices can become experts if they acquire the appropriate knowledge. 

\paragraph{Example~\ref{ex:aspect}}
An example of persons having shallow comprehension and shallow enabledness of the English present progressive tense could be beginning learners of English as a second language. Depending on their first language, they (may or) may not have prior knowledge of the function of the grammatical constructions. Their level of understanding of the present progressive tense could be characterized as being able to construct simple sentences using the present progressive tense of a number of verbs because these verbs have been introduced to them in the -ing form (shallow enabledness), but they are not aware that it is a marker of a specific tense and do not know when to use it (shallow comprehension).

\medskip
First, we want to describe the dimensions of understanding and the possibility of going through the field of shallow comprehension and shallow enabledness. To introduce changes in the model of understanding, we use the two dimensions of comprehension and enablement as parts of this model.

We can imagine that a novice in the field of shallow comprehension and shallow enabledness has still low comprehension of programming, does not yet understand the syntax of lines of code completely, does not yet know any concepts of iteration, and is not yet able to formulate it generally. However, improving the novice's understanding requires fundamental knowledge about the meaning of the code or the ability to interact with the algorithm to observe changes in the result. As is generally the case in all phases of an explanation and with regard to all forms of understanding, there are cognitive and motivational prerequisites of understanding at the beginning of an explanation. In the following, we focus in particular on prior knowledge and motivation for the situation of shallow comprehension and shallow enabledness. We would like to emphasize how an explanation comes about. In the situation that is usually at the beginning of a domain-specific explanation, prior knowledge, which is not only domain-specific, plays a crucial role. By looking at prior knowledge in this section in particular, we emphasize the importance of other, more general knowledge to characterize the basic knowledge in this situation. With the increase in understanding in the course of an explanation, does not only the explanandum change (cf. \citep {Miller2019})) but also the specific prior knowledge.  
Motivation is also a prerequisite for the (continuation of the) explanation in all the forms of understanding discussed in this paper. We present them especially in this section to show that motivation is essential to strive for an explanation as motivation is noteworthy for all actions (cf., \citep{heckhausen2018motivation}).

\paragraph{Prior Knowledge}

Prior knowledge is thought to influence co-construction between the explainer and the explainee \citep{GentileJamieson2021} because the explainer relies on the explainee's currently assumed prior knowledge to scaffold the explainee's understanding. However, a greater amount of prior knowledge on the part of both the explainer and the explainee does not necessarily lead to better explanations \citep{WilliamsLombrozo2010}.The influence of prior knowledge in explanations seems to be similar to its influence in learning scenarios: prior knowledge is not necessarily beneficial for the explanation process \citep{SimonsmeierFlaig2021}. It was found that the effect of explanation and the amount of prior knowledge on learning are neither independent nor subadditive \citep{WilliamsLombrozo2010}. Rather, they appear to have a nonlinear influence on learning. An explanation guides the focus in acquiring new knowledge, and prior knowledge helps to identify patterns and constraints. Not all prior knowledge is used as by explainers or explainees, only parts relevant to a task are activated \citep{DochyAlexander1995}. However, these ‘relevant’ parts are relative to the individual holding the prior knowledge \citep{AlbaHasher1983}. Thus, the personal relevant knowledge is not necessarily identical for the explainer and the explainee; neither is the assumed prior knowledge of the explainee on the side of the explainer necessarily correct. The assumed knowledge is part of a mental representation that the explainer has of his interaction partner, the partner model \citep{BrennanHanna2009}.

\paragraph{Assessment of Prior Knowledge}

The literature presents a variety of methods for the scientific assessment of prior knowledge. These include verbal methods such as ‘think-aloud’ tasks, verbal protocols and audit trails, and performance metrics on problem solving and troubleshooting or retention over time, as well as methods specific to human-computer interaction, system usage, user explanations of the system, and user prediction of system performance \citep{Sasse1991}. The simplest method is to query only the domain-specific prior knowledge \citep{BestOzuru2004}. However, this method discards most of the total prior knowledge.

An alternative approach is presented by \citet{ColeGaeth1986}, who use a combination of three different questionnaires to assess prior knowledge. The first questionnaire tests prior knowledge using true/false questions about domain-specific conceptual knowledge. In addition, they ask respondents to estimate their prior knowledge, such as \enquote{How well do you know \emph{X}?} The last questionnaire tests prior knowledge by asking about familiarity with the domain, e.g., \enquote{How often have you used \emph{X}?} In addition, the authors provide a questionnaire that asks about attitudes towards the domain, as they influence the explanation \citep{Tobias1994} as much as the actual knowledge the participants bring to an explanation.

In addition to prior knowledge, the motivation brought to an understanding process also influences its course. \citet{WadeKidd2019}, for example, have shown a bidirectional relationship between curiosity and learning. Therefore, the motivational prerequisites for comprehension will be discussed in the following. 

\paragraph{Motivational prerequisites}

Regarding explanations, the question is what motivates explainees to establish understanding of a specific domain as a goal. Their motivation can be decomposed into two aspects: The social motivation to follow the explanation and the motivation to understand. Consider the situation of students at school. Certainly, they are not always motivated by interest, achievement, or usefulness of the subject, but they know that it can be very useful (or even interesting) to follow the teacher's explanation.

To identify motivational prerequisites, a classical approach describes motivation as influenced by expectancy and value. Expectancy refers to \enquote{beliefs about how well they will do at the activity} (\citealt[p.~68]{WigfieldEccles2000}, cf. \citealt{EcclesWigfield2020}). The subjective value of understanding is divided into at least two parts: understanding can have (1) an intrinsic value of the activity of understanding, arising from an interest in the explanandum or a relation to the explainee in the social situation, or (2) understanding the explanation can be useful or valuable for achieving a goal, and thus has an extrinsic value that does not lie in the action itself. For example, understanding digital artifacts is useful or even required in the context of school, university, and the labor market \citep[e.g.][]{BallHuang2017}. Expectation as well as value depend on the previous experiences of the explainee \citep{BonanatiBuhl2021}. With the experience that it is difficult to follow an explanation or that digital artifacts are difficult to understand, one's self-efficacy to understand XAI-explanations is likely to be low. On the other hand, an adequate explanation increases the self-efficacy to use (digital) technology (\citealt{Weitz2021}; cf. technology acceptance models: \citealt{Davis1989}, \citealt{MarangunicGranic2014}). Both expectancy and value depend on the explainee's goals (\citealt{EcclesWigfield2020}, cf. \citealt{Covington2000}), such as learning the present progressive tense for a test in school or for living in a foreign country.

\subsection{Shallow comprehension, deep enabledness}
\label{sec:scde}

Although one can think of specific skills or cases of this form of understanding in its most extreme form, as we show in Examples 1 and 2 below, we argue in this section that there are not many cases of more complex actions following explanations that can be performed at the deepest level of enablement and still have shallow comprehension. Instead, we argue for actions that build on each other, that an increase in enabledness will lead to an increase in comprehension, and that these dimensions of understanding are intertwined. Thus, we also address the transitions between types and levels of understanding. We begin by describing how deeper enabledness can be achieved even when comprehension is rather shallow. In order to describe the processes, we also briefly refer to shallow enabledness, thus adding to the previous Section~\ref{sec:scse}.

\paragraph{Example \ref{ex:while}}
An example of people who have only a shallow comprehension but a deeper form of enabledness of a stopping criterion in an algorithm could be experienced self-taught programmers who lack formal training in logic and algorithms, but who can easily motivate their formulation of the stopping criterion for a while-loop (shallow comprehension). However, using their experience and possibly automated knowledge, they are able to get the conditional expressions right (deeper form of enabledness).

\paragraph{Example \ref{ex:aspect}}
An example of people with shallow comprehension and deep enabledness of the English present progressive tense would be native English speakers who intuitively know which form of the present tense to use and can construct it correctly (deep enabledness). However, if they were asked to explain their choice of the present progressive tense, they would not be able to do so (shallow comprehension).

\medskip
Besides the motivational prerequisites already explained in Section~\ref{sec:scse}, the access to external resources (e.g., tools such as web search, chatbots, FAQs, other people) may be relevant to be enabled to perform an action without necessarily having a conceptual comprehension of the task. One example is to program another while-loop and make a query to a chatbot (as an external resource) to support this task. In this sense, a \emph{deeper enabledness} is characterized by flexibility and adaptivity in the use of the external resources (e.g., novices vs. experts, see Section~\ref{sec:novice-expert}). The more flexibly an explainee uses the resources and adapts their use to different contextual environments, the deeper is the enabledness. Flexibility here means (i) various resources are known, and (ii) contextual factors (such as time restrictions, current (un)availability of an external resource etc.) are taken into account to choose the appropriate resources in a certain situation. Adaptivity means the ability to select the appropriate resources in novel contexts/in different contextual environments. In this respect, deep enabledness allows someone to use their skills in a wider range of applications, and without the need for resources being present in the immediate environment, because they can be simulated using internal mental representations, e.g., by memorization (see below). However, the deepest form of enabledness cannot be reached without any comprehension, since choosing what the best/most helpful resource is in a certain situation, or making a concrete query when using systems such as web search or chatbots, certainly presupposes some conceptual knowledge of the problem or task (cf. \citealt{GraesserPerson1994} on the relation between asking deep-level questions and deepening of conceptual knowledge).

In contrast, \emph{shallow enabledness} encompasses rather basic actions of an individual. This has, for example, been described for early infant development. \citet{Mandler2012} proposes that an infant might be able to identify an object as belonging to a category but this cognitive operation is based on perceptual information only and does not require an interpretation of spatiotemporal information. A basic action is embedded in the external representations itself and is externally driven (i.e., there is a perfect match between the physical disposition and the environment; see Identification/recognition in the listed abilities). In contrast to deeper enabledness, only a limited number of external resources are used and they are used context-dependent, in a given way (e.g., using the same query again) and cannot be transferred to another or a changed context. The \emph{transition from shallow to a deeper enabledness} comes by experience, i.e., a higher degree of flexibility and adaptivity is achieved by richer experience (through repetition in the same or only slightly changed contexts) and a far transfer of skill \citep{Garder1999}. Enabledness without deeper comprehension may also be achieved by memorizing a number of steps, procedures or facts (rote learning) without understanding the concepts behind it. However, this approach to enabledness, that \citet[p.~9]{BoalerGreeno2000} call a \enquote{ritualistic act of knowledge reproduction}, may hinder progress in deepening comprehension because explainees may be less engaged with the underlying concepts. Deeper comprehension, however, requires \enquote{thinking practices} \citep[p.~9, for mathematical learning]{BoalerGreeno2000}. Therefore, we assume that deeper enabledness as we have defined it, including a flexible use in different contexts, cannot be accomplished by memorization or rote learning.

To define \emph{comprehension}, among others we made use of the taxonomies proposed by Bloom \citep{KrathwohlBloom1969} and Krathwohl \citep{AndersonKrathwohl2001,Krathwohl2002} to classify learning objectives. They describe different cognitive processes that a learner is able to perform within a learning process. However, originally there is no clear distinction between enabledness and comprehension in these learning taxonomies: Bloom's taxonomy \citep{KrathwohlBloom1969} starts with early processing mechanisms that we ascribe to enabledness but not comprehension, whereas Krathwohl's taxonomy \citep{AndersonKrathwohl2001,Krathwohl2002} seems to require interpretation and memory from the beginning. To make the early steps of understanding clear, we will therefore refer to Bloom’s taxonomy. Overall, we recognize that there is a progression from being bound to perceptual information or the context to decontextualized knowledge that is more flexible and adaptive. We view the boundedness to external resources not only to be more shallow in terms of enabledness (see above) but also in terms of comprehension. Therefore, the abilities described in the first two stages on Bloom's taxonomy seem to belong to shallow comprehension/shallow enabledness (see Section~\ref{sec:scse}). It is difficult, however, to make a clear cut between comprehension and enabledness because the different forms of comprehension only become observable by abilities, and therefore by forms of enabledness.

\begin{description}

    \item[Identification/recognition]
    \citet{KrathwohlBloom1969} identified the following abilities as reflecting the first steps of comprehension: Repeating, listing, labeling, completing, and selecting. We propose that all of them afford a more or less prepared context (i.e., external resources) in which the knowledge is triggered to be applied. According to \citet{Mandler2012}, the operations at this level may be based on perceptual information only and do not require an interpretation of the context. For the other abilities, based on the rich context, only little knowledge or comprehension is required to achieve an act (e.g., to label something or to fill something out).
    
    \item[Contextualization: Connection to the obvious/present]
    The second level, called "understanding" in Krathwohl's taxonomy \citep{AndersonKrathwohl2001}, represents, among others, the abilities of interpreting, exemplifying, and summarizing. We propose that they require some knowledge to be connected to another entity that is obvious or present in the environment (i.e., contextualization), thus requiring a deeper comprehension than in the first stage. Contextualization was recognized as a form of interpretation of an activity based on cues from the obvious/present and related to the process of inferencing, which itself depends on the context of its occurrence \citep[p.~30]{Auer1992}. For our example~\ref{ex:while}, a reasonable explanation for why the first result is ‘99 bottles …’ is that \texttt{counter} is initialized with ‘99’. This explanation simply accounts for what can be given by referring to the present context, without having to add anything to it. A deeper reason goes beyond the obvious and offers a connection to what cannot be perceived, such as what the output would be if the code in line 1 or 6 were changed in a certain way. Making connections is important in the learning process \citep{GentnerLoewenstein2003}. 

    \item[Applying (transitional step towards deep comprehension)]
    We see the third step of comprehension at the intersection of shallow and deep, i.e., as intermediate, because it requires applying the knowledge to a context that is not necessarily prepared for it (i.e., the context does not trigger a specific application). The application of knowledge always leads to deeper enabledness.

\end{description}

All of the three above-mentioned steps may be time-dependent -- comparable to cognitive mechanisms of mapping discussed for word-learning in infancy \citep{Wojcik2013}, i.e., fast-mapping (short-term; less demanding) and slow-mapping (requiring long-term memory; more demanding), or to different forms of cognitive engagement of students \citep[p.~221]{ChiWylie2014}. \citet{ChiWylie2014} differentiate between recall (what we see as shallow comprehension), apply (at the intersection from shallow to deep comprehension), transfer and co-create (deep comprehension). 

In the following, we argue that comprehension and enabledness are intertwined. We presume that most types of deep enabledness would build on some kind of comprehension, and, vice versa, deeper enabledness might lead to increased comprehension: (i) comprehension changes depending on enabledness and vice versa, and (ii) comprehension, even shallow, only becomes observable in physical and cognitive abilities that are also signs of enabledness (cf. Section~\ref{sec:intertwinement}). Cases of extreme shallow comprehension together with extreme deep enabledness, or cases of shallow comprehension and intermediate enabledness are therefore hard to be defined and would contradict all propositions made above. (iii) There is neurological evidence on language production, according to which comprehension and enabledness cannot be torn apart at the cognitive level because of shared underlying neurological representations, at least from a functional perspective \citep[e.g.,][]{PickeringGarrod2013}.

In terms of \emph{assessability of comprehension}, the question is how verbalization is related to the degree of comprehension. This question arises based on studies on gesture use and learning (cf. extensive work by Goldin-Meadow and colleagues, e.g., \citealt{Goldin-Meadow2017} for an overview): If comprehension is shallow, a person may not be able to verbalize knowledge, i.e., the knowledge is more implicit. However, Goldin-Meadow and colleagues have shown that other (communicative) behavior such as a person's gestures can indicate some initial comprehension: even though some learners failed to solve some mathematical problems and were unable to explain their solution, their co-speech gestures conveyed semantic information that was relevant to correctly solve the task but that was not verbalized. Thus, it was assessable nonverbally. The finding that these learners were more likely to benefit from further explanations was interpreted in terms of them having implicit conceptual knowledge. Research on implicit conceptual knowledge has further shown that participants were able to make judgments on whether a problem-solving task is likely to be solvable or not, even when they were not able to provide a correct solution themselves \citep{BowersRegehr1990}. When comprehension becomes deeper, it becomes explicit, that is, it can be verbalized, and is therefore assessable by asking questions.

\paragraph{Intertwinement of shallow comprehension and deep enabledness}

\begin{figure}
    \centering
    \includegraphics[width=\columnwidth]{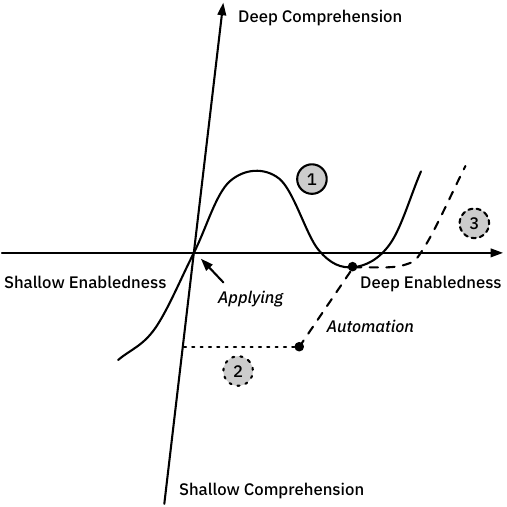}
    \caption{Three cases illustrating the strong intertwinement between comprehension and enabledness.}
    \label{fig:three-cases}
\end{figure}

As mentioned above, we assume a strong intertwinement between comprehension and enabledness. However, it should not be understood as a linear increasing relationship. Instead, the following three cases, see Figure~\ref{fig:three-cases}, display the dynamic relation between both forms of understanding. 

The first case is a process at the transition from intermediate to deeper enabledness with alternating more shallow and deeper comprehension (solid line in Figure~\ref{fig:three-cases}), that may often occur during automation, e.g., in language (L1) acquisition. If something becomes automatized, enabledness becomes deeper, but comprehension may decrease \citep{Anderson1996, BoalerGreeno2000}, as during the process of automation conceptual knowledge becomes compiled, and therefore less accessible (see Section~\ref{sec:aspect-of-understanding}). An example could be automation demonstrated by highly proficient bilinguals in terms of their language switching abilities in everyday life \citep{GreenAbutailebi2013}. Research suggests that initially, during the early years of second language acquisition, non-selective language activation occurs. However, as both languages become dominant, the selection rules become automatic. This leads to automatic concept switching, where bilinguals become unaware of the switching rules and rely more on external resources like social cues and a partner's fluency \citep{GreenAbutailebi2013}.

In the second case (dotted line in Figure~\ref{fig:three-cases}), comprehension stays at a rather intermediate level although enabling increases. This is only possible if multiple external resources are available,  flexibly used (see definition of deep enabledness) and  combined by a person (e.g., to solve a programming problem, a colleague can serve as an external resource, or in other situations, possibly due to time restrictions, a web search could be another external resource). However, we think that this scenario allows deeper enabledness only to a certain extent because deep enabledness requires deep comprehension. An example representing this could be a situation in which a lay user without any programming skills can run a script written in a certain programming language. With already available commands that can be placed on the correct position of the programming console, the lay user may manage to complete the task successfully, and may even be enabled to replicate this in a slightly changed context. Thus, their enabledness becomes deeper (though only to a certain extent), but comprehension (e.g., programming syntax and meaning) would remain at a constant, more shallow level if no further explanations are given.

The third case (dashed line in Figure~\ref{fig:three-cases}) is linked to the second one. A progress in enabledness may result in a lagged, nonlinear increase in comprehension following further explanations, that is, the progression in both forms of understanding is stepwise, similar to the idea of bootstrapping from developmental research (\citealt{MorganDemuth2014}; cf. Section~\ref{sec:intertwinement}). This is based on research on automated processes in second language (L2)learning \citep[cf.][]{HaoOthman2021}. Although L2-learners might encounter difficulties in the process of generating new knowledge, for example the present progressive tense, this process may be supported by establishing relations to previously acquired, related conceptual knowledge, such as the general distinction between present and past, as well as knowledge expressed in identification of (in-)consistent syntactic constructions of the L2 based on their L1 knowledge (cf. \citealt{Mandler2012}; \citealt[p.~7]{FiorellaMayer2015}). At this stage, comprehension remains behind the intermediate level, but enabledness may increase progressively if L2-learners apply their previous conceptual knowledge by doing and repeating related exercises, e.g., filling missing auxiliaries ‘to be’ or missing {-ing} suffixes. Learners may also rely on external resources such as textbooks, resources for language practice, or a human tutor to complete the tasks. Learning the different application contexts of the English present progressive tense would enable learners to use the present progressive without recalling basic concepts and to apply it to various similar cases in L2-learning.

To conclude, in our view and because of the intertwinement there are only limited cases of really shallow comprehension and deep enabledness, and there are no real life scenarios for the extreme, i.e., shallow comprehension with intermediate-to-deep enabledness.

\paragraph{Assessment}

If we follow Krathwohl's taxonomy, for which we have argued that the first and second steps require rather shallow comprehension, then tasks such as summarizing can be used to assess shallow comprehension. To assess different levels of enabledness, different tasks should be successfully accomplished with the support of external resources (see above) that are either very similar to those present during the explanation (more shallow/intermediate enabledness) or that are modified to assess whether the knowledge can be flexibly applied to novel contexts or other use cases (deeper level of enabledness) within the specific domain of what was explained.

Regarding automation, i.e., how efficient an explainee is, different measures could be used, e.g., processing time, or number of errors in relation to total performance. Since we have defined automation as compiled knowledge that may be less accessible to verbalization, the assessment could also consist of two parts, a task and a questionnaire: if the explainee is able to successfully accomplish the task but is unable to explain why or how it was done, comprehension is likely to be shallower. However, this may also be an indicator of transitional conceptual knowledge \citep{Goldin-MeadowAlibali1993}.

\subsection{Deep comprehension, shallow enabledness}
\label{sec:dcse}

With respect to the intertwinement discussed in Section~\ref{sec:scde}, deep comprehension combined with shallow enabledness is a rare case. However, it is a persistent problem, e.g., in educational contexts, which are often ultimately aimed at in educational settings. It can, and often does, occur without a translation into “knowing how” as an indicator of shallow enabledness. An example of lacking conditional or conditionalized knowledge is a student with knowledge about learning strategies who, in a school situation, does not recognize that a problem is about learning strategies (and not about motivation, etc.).  Another example of a lack of compilation is a person who has a lot of knowledge about hand spinning (e.g., types and functions of spinning wheels, properties of fibers, ideas of twist, and the basic knowledge of how to spin and ply them into a stable thread). Without the procedural knowledge of how much twist the yarn should have, how fast to draw out the fiber, how to stop the wheel, and how to counteract errors, this deep comprehension leads only to shallow enabledness. This form of knowledge can therefore also be characterized as the combination of existing conceptual or even theoretical knowledge with a lack of practical application, e.g., due to a lack of practice.

One reason for deep understanding and shallow enabling is what has been called ‘inert knowledge’ \citep[p.~5]{Whitehead1929}, although it was more concerned with inert (scientific) ideas than inert knowledge of individuals. More recent psychological research has demonstrated the prevalence and relevance of this problem and suggested several explanations for it that are different in nature and underscore its complexity \citep{RenklMandl1996}. The relationship between deep understanding and shallow enabledness is here conceptualized in a very specific way.
According to meta-process explanations, the knowledge would be applicable, i.e., would result in enabledness, but external prerequisites such as motivation or the perception of benefits of applying the knowledge are missing. Also, persons might lack information about when or where to apply knowledge, which would also show up as shallow enabledness.
Other explanations focus on a deficiency in the structure of knowledge. Important ones that could also explain the combination of deep understanding and shallow explanations are a lack of compilation and a lack conditionalized knowledge. Declarative knowledge has to be compiled in order to be used, and if compilation is insufficient or if a person lacks knowledge about conditions in which knowledge can or cannot be applied, shallow enabledness results.
Finally, it should be noted that theories of situated cognition do not regard inert knowledge as a problem, but rather as due to the fact that knowledge is bound to certain contexts and situations and their specific conditions \citep[e.g.,][]{Greeno1998}.

\paragraph{Example~\ref{ex:while}}
Our example of a programmer who fully grasps the stopping criterion of the algorithm, but is not able to formulate it recursively might fit this explanation.

\paragraph{Example~\ref{ex:aspect}}
Our example of a second language learner who knows the concepts of the present progressive and how to build it, but cannot formulate correct utterances spontaneously in a dialogue might also fit this explanation.

\subsection{Deep comprehension, deep enabledness}
\label{sec:dcde}

Following the definitions of comprehension and enabledness, the combination of deep comprehension and deep enabledness is an important prerequisite for what is often referred to as ‘agency’. In this section, after presenting what deep comprehension and deep enabledness would be in the case of the two examples, we elaborate on the theoretical background of the term agency.

\paragraph{Example~\ref{ex:while}}
In general, we reason that programmers have ‘agency’ if they can identify and formulate the stopping criterion effortlessly and correctly. For implementing the algorithm in a certain context, they apply theoretical knowledge about iterative algorithms and stopping criteria and consider the demands defined by the context (comprehension). Furthermore, they apply this knowledge when defining the stopping criterion in a specific way, which is implied by the programming language that is used (enabledness). Only if they do both and can flexibly combine their understanding of the algorithm with their abilities to write the program in a programming language, a performance of high understanding, and thus agency, can be observed.

\paragraph{Example~\ref{ex:aspect}}
Agency in the case of languages can be described as communicating competently and appropriately in a foreign language and being able to give reasons for formulations. For this, it is not sufficient to be a competent speaker; you also need to be able to reflect upon linguistic choices with reference to a grammar system. A speaker competent in English would know that the progressive is generally in the form auxiliary verb be + verb ending in -ing, and they would be able to adhere to these rules in writing and speech as well, without actively thinking about it too much. However, scholars with a large body of knowledge and practical experience can additionally elaborate why in certain contexts the present progressive tense is the correct choice, as well as, for example, describe common errors of native speakers. Simply knowing a handful of contexts and use cases of the present progressive by heart would not be sufficient. Instead, scholars can flexibly use and apply their knowledge to come up with novel contexts, depending on the situation (e.g., when asked a question about the rules by English language novices). The fact that scholars have meta-knowledge and can flexibly use the progressive form (enabledness, knowing-how) in combination with what they know (comprehension, knowing-that) shows their understanding and agency.

\paragraph{Definitions of agency} 
Agency is defined in various ways, which is apparent as the term itself is often combined with another term to form an open compounds part of a specific phrase such as, for example, ‘digital agency’ (\citealt[p.~736]{ArnoldClarke2013}; see, e.g., \citealt{PasseyShonfeld2018}). There is the need for scholars to provide a precise definition of the term \citep[cf.][p.~130]{Ahearn2001} and the vagueness of the term is criticized, too \citep{EmirbayerMische1998}. Different definitions help to illustrate the problem: \citet{PasseyShonfeld2018} define ‘digital agency’ quite broadly as \enquote{Digital Agency (DA)—consisting of digital competence, digital confidence, and digital accountability—is the individual’s ability to control and adapt to a digital world}. \citep{Couldry2014}, in his inaugural lecture, was more explicit: {First, agency, by which I mean not brute acts (of clicking on this button, pressing ‘like’ to this post) but (following Weber) the longer processes of action based on reflection, giving an account of what one has done, even more basically, making sense of the world so as to act within it}. \citet[p.~289f]{Kallinikos2002} describes how technology may influence agency \enquote{by inviting specific courses of action} that may be shaped to a certain degree (‘malleability’) and, that agency is \enquote{captured by the distinctive ways by which a technology invites people to frame a delimited domain of tasks or activities and organize their execution}. \citet[p.~751]{ArnoldClarke2013} define student agency as \enquote{[t]he discursive practice of positioning oneself or being positioned as responsible}. \citet[p.~963]{EmirbayerMische1998} follow the pragmatic tradition of Bourdieu and Giddens, who define \enquote{[…] human agency as habitual, repetitive, and taken for granted—a view shared by ethnomethodologists, new institutionalists in organizational theory, and many others}. According to \citet[p.~300]{Sharma2008} agency is a \enquote{contingently emergent feature of situated local action}. Thus, humans do not just have agency, but they can show agency through their actions. \citet{SiryLang2010} describe agency as a feature a person can acquire and develop. This means humans may acquire agency during childhood and education. \citet[p.~891]{Basu2008} describes agency as purposefully examining and reflecting the actions of one person; the foundation of this acting is the beliefs and goals one person has. She says: \enquote{[…] I describe agency as purposefully considering and enacting both small- and large-scale change in personal and community domains, based on one's beliefs and goals}. \citet[p.~453]{Duranti2005} defines agency more broadly, as something entities are having. His working definition defines agency as a property of entities to (i) have a certain degree of control over their behavior, (ii) whose actions have a certain effect on other entities, and (iii) whose actions can be evaluated for example for responsibility.

\paragraph{Agency in the context of understanding}
A mid to high degree of enabledness combined with a mid to high degree of comprehension are needed for agency. If both are pronounced, an individual is capable of actively manipulating and controlling various situations flexibly, by using what they know to reach individual goals. A high degree of comprehension and enabling is agency as it is a prerequisite for various properties mentioned in definitions by \citet{PasseyShonfeld2018} (\enquote{control and adapt}), \citet{Basu2008} (\enquote{purposefully considering and enacting both small- and large-scale change in personal and community domains, based on one's beliefs and goals}), \citet{Duranti2005} (having a certain degree of control over one's own behavior, having an effect on other entities through one's behaviour) and \citet{Sharma2008} (\enquote{a contingently emergent feature of situated local action” perceivable within actions}). 

The question of how one would assess understanding in specific contexts and therefore, implicitly, how agency arises, is challenging is the fact that our concept of understanding remains a latent construct that is not trivial to define. Especially mental operations like connecting new facts cannot easily be measured. In other words, understanding cannot be assessed by asking to recall facts. However, it can be observed when people are engaged in unknown situations that require the utilization of previously acquired knowledge and a person's enablement. In these situations, agency results in a flexible solution to the given task. This approach adheres to Perkins’ idea of measuring the ‘performance of understanding’ \citep{Perkins1993, Perkins1998}. Perkins recognizes \enquote{understanding through a flexible performance criterion} that \enquote{shows its face when people can think and act flexibly around what they know}. In contrast to this, a signal of lack of understanding is, \enquote{when a learner cannot go beyond rote and routine thought and action} \citep[p.~42]{Perkins1998}.

\section{Dynamics of Understanding}
\label{sec:dynamics-of-understanding}

Understanding is a process that unfolds over time when an explanation is successful \citep{Keil2006}. In this section we offer a conceptualization of the dynamics of understanding over time. We focus on two distinct aspects that were mentioned in passing above: differences between novices and experts, and the intertwining of comprehension and enabledness in the transition from a lower to a higher form of understanding.

\subsection{Novices and experts}
\label{sec:novice-expert}

Using the heuristics of our four-field diagram, an explainee with shallow comprehension and shallow enabledness is considered a novice, while a deeply enabled explainee with deep comprehension is considered an expert. The purpose of this section is to outline some fundamental differences between novices and experts with respect to the two types of knowledge.

First, novices and experts are assumed to differ in terms of procedural and conceptual knowledge. Experts were found to categorize problems by applying abstract solution procedures (e.g., major physical principles), whereas novices build their problem categorization on \enquote{the entities contained in the problem statement} \citep[p.~15]{ChiFeltovich1981}. In terms of problem representation, experts translate features of the problem, thereby activating abstract category schemes for a problem type that are organized, for example, by universal laws or principles of nature. In contrast, novices organize their representation of a problem by schemata for object categories, thus staying on more concrete levels when formulating the problem. However, a novice can become an expert in a field by acquiring the relevant domain knowledge. The concept of the beginner is therefore fundamentally development-oriented.

Second, while domain knowledge is one aspect in which novices and experts undoubtedly differ, they are also likely to differ in terms of experience, e.g., in a communicative practice. A study investigating novice tutors' abilities to assess and monitor students' knowledge was found to be profoundly inaccurate \citep{ChiSiler2004}.

In our attempt to go beyond classical cognitive psychology, which distinguishes between experts and novices by focusing on (domain) knowledge, we would like to enrich the pair of terms by bringing in constructive learning theory, which considers context orientation and situatedness as crucial. The focus here is on the interactive processes of how experts guide novices into ways of participating in a practice after a shared goal has been established in a given situation \citep{LaveWenger1991,Rogoff1990}. 
For example, children are often referred to as novices because they are less knowledgeable. In conversations, the more knowledgeable participant (i.e., usually the adult) provides interactive support to help the child perform a aspired task \citep{WoodBruner1976}. \citet{Goodwin2013} broadens the perspective even further by explicitly including the body in the conceptual pair, thereby including multimodal aspects of interaction. He points out that in order to become competent, novices need to train both their bodies and the perceptually relevant structures of the objects or surroundings to be scrutinized in order to become skilled at performing a particular practice.
In this sense, \citet{Goodwin2013} provides an anthropological view of embodied cognition that sees body and mind as inextricably intertwined. Understanding is therefore displayed through correct embodied behavior, and teachers could use displays of embodied behavior to assess students' knowledge \citep[see also][for perceptual structure provided by caregivers]{Zukow-Goldring1996}. \citet{HindmarshReynolds2011} shows how supervisors of dental students rely not so much on talk but on bodily behavior to assess their students' ongoing understanding: even though students may claim to understand a procedure, the way they move their bodies to look into a patient's mouth may actually reveal their lack of understanding of the matter at hand. 

Including embodiment in the conceptualization of understanding adds a new perspective because it may provide a more complete view of understanding than simply focusing on internal domains of knowledge. Novices would then differ from experts not only in terms of conceptualization and problem-solving procedures with respect to a particular domain, but also in that they would not have the embodied knowledge of how to perform a task \citep{Zukow-Goldring1996}.
It has been pointed out that knowledge and its asymmetries are relevant in distinguishing between experts and novices. The next section looks at different types of knowledge in this context.

\subsection{Intertwinement of comprehension and enabledness: Transitions between the forms}
\label{sec:intertwinement}

The idea for the intertwinement of comprehension and enabledness (see Section~\ref{sec:scde}) is related to the concept of ‘bootstrapping’ which might be an underlying mechanism facilitating the transgression from one part of the four-field diagram to another. Bootstrapping is assumed as being a learning mechanism for linguistic development by which growing abilities on one linguistic level, e.g., phonology, supports the progression on other linguistic levels such as syntax or semantics \citep{MorganDemuth2014}.

Applying this to the different forms of understanding in our four-field scheme, comprehension can be bootstrapped by enabledness because comprehension assumably deepens when learning to apply a knowledge domain to a novel context/situation. Vice versa, enabledness deepens with comprehension, i.e., learning more about covert mechanisms of a domain. As an example, enabledness in programming a while-loop will improve when the person learns more about programming languages. Thus, bootstrapping could explain the transitions between different forms and levels of understanding. 

Example~\ref{ex:while} can be used to illustrate this mechanism in the early stages of understanding (see Section~\ref{sec:scse}). The challenge for a novice is to understand the stopping criterion of an iterative algorithm through explanation, or to devise a strategy for understanding the basic theory of a while-loop in a programming language. To describe the states of the model of a novice’s understanding in this example, we are looking at the transitions to other forms of understanding from the field of shallow comprehension and shallow enabledness:

\begin{description}

    \item[(1) Shallow comprehension, shallow enabledness] The starting point depends on the novice's prior knowledge. A novice with basic computer science knowledge will start by modifying sample code functions and examining the results to improve their comprehension of the functions and their meanings. A novice with no computer science knowledge, but with a higher conceptual knowledge about the syntax of an iterative function, will start to reproduce the results manually on paper. In both cases, interaction with the code will increase the model of understanding in an exploratory way.

    \item[(2) Shallow comprehension, increasing enabledness] A nov\-ice with less conceptual knowledge about a while-loop, but more knowledge about how to interact with code, can try out the code example. The process of reproducing the meaning of the code will increase (i.e., bootstrap) the understanding of the syntax and meaning of an iterative function. However, the lack of comprehension of the theoretical background and the repetition of such a function (e.g., the stopping criterion) can also lead to irritations and misunderstandings. Thus, an incorrect model for the theory of a while-loop could be built up.

    \item[(3) Increasing comprehension, shallow enabledness] Increased comprehension can lead to an understanding of the code syntax and the overall goal of the function. For example, a novice knows in theory that a while-loop needs a stop criterion to let the algorithms finish a computation. But with shallow enabledness, the novice may not be able to modify the code independently without generating errors. In addition, the novice may not be able to apply their comprehension to new contexts outside this specific example.

    \item[(4) Increasing comprehension, increasing enabledness] A nov\-ice with an increased comprehension knows the meaning of the symbols in the code and understands the basic syntax. A novice might be able to understand the theory of a while-loop, mainly by interpreting the iterative repetitions in a given example, and reproduce the expected results outside of that example.

\end{description}

These transitions can, for example, be classified in the field of shallow comprehension and shallow enabledness as follows: (1) The initial prior knowledge of a novice, (2) the process of how to access conceptual knowledge without known theories, (3) understanding the general theories without being able to apply them, (4) building up a strategy to increase conceptual knowledge by applying less or no known theories.

\section{Forms of Understanding in a Classic XAI Scenario}
\label{sec:classic-xai-example}

In Section~\ref{sec:introduction}, we argued that our four-field model of forms of understanding is applicable to explanations in both human interaction and human-explaining AI interaction. The explaining AI, however, was assumed to be a new form of a ‘social‘ explainable AI which is co-constructive and interactive \citep{RohlfingCimiano2021, Rohlfing2025socialxai}. This raises the question of whether the model would, in principle, also be useful in ‘classic’ XAI scenarios, e.g., where explanations take the form of counterfactuals. A problem with such XAI scenarios is that users who are interested in explanations of outcomes produced by AI systems already require expertise in interpreting the form of explanation that is provided by a system. For example, they need to be able to work with LIME explanations \citep{Ribeiro2016KDD}, or know how to formulate contrastive questions and interpret counterfactual explanations \citep{Miller2021, Miller2023} -- and how these forms of explanation can be used to improve understanding of outcomes.

In this section we show that our four-field model of understanding can also be usefully applied to such scenarios by applying the model to the classic XAI scenario in which an automated system decides whether to grant a client a bank loan and under what conditions. Let's assume that the client uses a bank's website that offers an interface for customers interested in loans and that this interface offers counterfactual explanations. We assume that this client has relevant XAI experience, i.e., they know how to interpret counterfactual explanations and ask contrastive questions, but have no domain expertise. In other words, they do not know much about how bank loans are calculated and decided upon.

When the client starts to interact with the system, they can thus be assumed to have shallow comprehension and shallow enabledness on how the bank will make the decision. First, the system requests relevant information from the client, such as their demographics, income and credit score. It also asks for information on the loan they are seeking, e.g., the loan amount, term and interest rate, and how it will be used. By providing all this information, the client may already gain some comprehension that the decision and conditions are probably based on the specific parameters they provided. The system initially rejects the client's request, prompting them to engage with the XAI component and ask contrastive questions about their income, the loan amount, loan term, and its intended use. This deepens the client's comprehension of the decision procedure, e.g., that the input variables are not necessarily independent, and that making smaller changes to multiple variables may be more effective than making larger changes to a single variable in order to achieve a different outcome.

By experimenting with the input variables and posing further contrastive questions, the client can explore the system's decision-making process in more depth, deepening their enabledness to predict the decision based on changes to their input. They can now better understand which types of loans they are eligible for and which personal factors they are in principle able to change to qualify for the loan they initially aimed for (see, e.g., \citealt{MindlinRobrecht2024} for an experimental investigation in measuring users' enabledness to predict income using a dialogue-based XAI system). This gives them agency and empowers them to make informed decisions about their financial situation. 

It is important to note that, in classic XAI scenarios, the client's ability to achieve understanding depends on two prerequisites:
    (1) knowledge of how to interpret the forms of explanation provided by the system (e.g., formulating contrastive questions and interpreting counterfactual explanations); and
    (2) knowledge of how to systematically make use of the system's forms of explanation to actually facilitate understanding of the its outcomes.
As a result, the responsibility for achieving understanding rests entirely with the client. An interactive and co-constructive social XAI system, in contrast, would adapt its explanation and scaffold the client's understanding \citep{RohlfingCimiano2021}, making it a more collaborative process.

\section{Conclusion}
\label{sec:discussion-conclusion}

As shown in the previous sections, discussing the cognitive side of XAI, it is important to consider types and forms of human understanding of explanations depending on their prior knowledge, motivation, and goals. These aspects are increasingly considered in the XAI literature \citep{Miller2019, Miller2023, SpeithCrook2024, RohlfingCimiano2021, 
Rohlfing2025socialxai}.

We suggested aspects and forms of understanding which may occur in explanations between humans, but also in XAI-explanations for human users. Considering the forms given in our four-field diagram (Figure~\ref{fig:types-and-levels-of-understanding}) underlines that explainers as well as explainees pursue many different goals: to be enabled to do something, to comprehend something, or to achieve full agency. This agency needs -- as we assume -- both, deep comprehension and deep enabledness. There may, however, also be situations, in which humans only strive for shallow enabledness and/or shallow comprehension. The development of XAI technology has to consider these possibilities as well.

Generally, understanding is a journey from shallow to deep comprehension and enabledness, which results in both forms being highly intertwined and in need of being considered. In the end, it is not possible to reach deep enabledness without comprehension. For XAI, it follows that the amount of information required needs to be negotiated between the explainer (which could be an XAI system) and their explainee. As an important background of our contribution, \citet{RohlfingCimiano2021} conceptualize explaining as a social practice in which explainers and explainees are equally involved in (and responsible for) co-constructing the explanation and its understanding.

We consider these conceptualizations to be a concise theoretical structure for an interdisciplinary discussion and multi-perspectival development and evaluation of (social) XAI technology, and explaining and understanding more generally. Two issues that we have not addressed above but that are highly relevant for explainability are (1) the need to consider that humans often have a subjective understanding that does not fully align with their actual/objective understanding and (2) that our cognitive approach of conceptualizing understanding neglects that explanations always take place in a social context. We close this paper by discussing these two aspects.

\subsection{Subjective vs. objective understanding}

‘Subjective’ understanding in explanatory interactions relates to how well explainees think they have understood an explanation, whereas ‘objective’ understanding describes the \emph{actual} understanding explainees have of the explanans. Discrepancies between an explainee’s subjective claims and their objective understanding are to be expected (e.g., due to misconceptions or unknown unknowns). Concerning the forms of understanding proposed in this paper, an explainee's subjective comprehension is likely more prone to be misaligned to their objective comprehension than is the case for enabledness. The illustration of the assessment of the different forms of understanding in Section~\ref{sec:forms-of-understanding} shows that research makes use of subjective self-report measures of understanding as well as objective testing following learning taxonomies. 

The distinction between subjective and objective understanding is also important in relation to biased responses -- as mentioned above. As long as we want to measure subjective understanding, we have to take into account that the answers may be biased in some way. This is not a problem in itself, but it becomes a problem as soon as the bias is caused by the question posed by an explainer, which is more likely to happen the more complex and detailed the question becomes.

In XAI, methods and interactive explanation systems that aim at co-constructing the human user's understanding, relying on monitoring and adaptation processes for tailoring explanations to users' needs \citep{RohlfingCimiano2021}, need to be aware of the difference between subjective and objective understanding. Depending on the mode of interaction, evidence of understanding that users provide can simply be subjective \emph{claims} of understanding, (e.g., through the production of verbal acknowledgments signals or indications of satisfaction in surveys) or \emph{demonstrations} of understanding that are more revealing of their actual understanding (e.g., being able to execute specific actions or asking follow-up questions that are contingent on understanding; cf. \citealt{Koole2010}). XAI systems need to be able to identify these different types of evidence, weigh them accordingly, and assess them critically when monitoring and adapting to their human interaction partners.

\subsection{Influences brought by society}

So far, the question of the potential impact of social processes on explanations and related processes of understanding has been excluded. However, how explainable AI -- or any other new technology -- is reflected upon in society has implications for the understanding of the individuals interacting with it, as humans constitute social reality by interpreting and creating their world and its objects in interactions \citep{Rosenthal2018}. Since the purpose of co-constructing Explainable Artificial Intelligence (XAI) is to support both machines and humans in their communication with each other, the understanding capabilities of both should be considered. The considerations relate to social constructionist approaches, according to which successful social acceptance, adoption and understanding of newly introduced technologies are highly dependent on the meaning co-constructed in social interactions, i.e., when human (or non-human) agents or group members interact. In this context, group membership (e.g., constituted by gender, class, race, age, socioeconomic status) plays a key role in the process of understanding technological novelty and thus in explaining it. Group membership \enquote{determines what is relevant and what is not} \citep[p.~361]{Kronberger2015}. In other words, goals or relationships may be different for different social groups. 

The social constructivist view of human interaction is also relevant to human-machine interaction, and even precedes it. \citet{EhsanRiedl2020}  emphasize the importance of a human-centered approach to XAI. Following their ‘reflective sociotechnical approach’, we propose a concept of reflective understanding that critically reflects the dominant narratives as well as the explicit and implicit assumptions and practices of the field. Moreover, it is value-sensitive to researchers, designers, and users of AI, including to those who tend to be excluded from the conversation and are denied access to the technology \citep[p.~461]{EhsanRiedl2020}. In other words, in order to understand how and why people act the way they do (e.g., adoption or non-adoption of a particular new technology) social meanings, interactions and interpretations cannot be ignored. Accordingly, explanations must also address how people interpret their world and how they interactively create this world. Not only the perspectives and stocks of knowledge that we are aware of should be of interest when designing an explanation, but also implicit knowledge and the interactive creation of meanings that people (users) are usually not aware of \citep{Rosenthal2018}.

In this paper, we have outlined a model of understanding that prioritizes the individual cognitive aspects of understanding. However, we propose to add a social dimension that inevitably informs cognitive processes in ways that are conducive to a conceptualization of co-constructive explanatory processes. Thus, we argue for a broad notion of understanding that goes beyond a cognitive model that conceives of comprehension and enabledness as discrete processes occurring in the minds of explainees, and argue that aspects such as context, interaction, or group membership, as well as broader socio-technical relations and references, should also be considered in the development of XAI technology. 

Applying our concept of agency (see Section~\ref{sec:dcde}), a high degree of comprehension and enabledness is characterized by healthy distrust \citep{VisserPeters2025}, deep enablement, knowledge of the strengths and power, but also of the limitations, biases, and potential for and actualization of oppression and exclusion that AI technology entails. In addition, agency means being empowered to overcome socially and technologically consolidated marginalization/discrimination that may be related to group membership. Thus, in order to support agency, explainers (human or XAI) should take into account explainees' group-related inequalities and differences, as well as varying access to resources and positions/roles in socio-technical systems, because group affiliation \enquote{determines what is relevant and what is not} \citep[p.~361]{Kronberger2015}. Indeed, we see this as inevitable for AI development because a conceptualization of human understanding of XAI-explanations without context is insufficient, and concepts of human understanding should be central to the research and design of XAI methods and technology so that they can do justice to the humans they aim to empower.

\section*{Author Contributions}

\noindent
\textbf{Helen Beierling}: 
    Writing – Original Draft.
\textbf{Heike M. Buhl}: 
    Supervision, 
    Conceptualization, 
    Writing – Original Draft, 
    Writing - Review \& Editing.
\textbf{Hendrik Buschmeier}: 
    Supervision, 
    Conceptualization, 
    Writing – Original Draft, 
    Writing - Review \& Editing.
\textbf{Josephine Fisher}: 
    Writing – Original Draft.
\textbf{Angela Grimminger}: 
    Conceptualization, 
    Writing – Original Draft.
\textbf{André Groß}: 
    Writing – Original Draft.
\textbf{Ilona Horwath}: 
    Conceptualization, 
    Writing – Original Draft.
\textbf{Friederike Kern}: 
    Supervision, 
    Conceptualization, 
    Writing – Original Draft, 
    Writing – Review \& Editing.
\textbf{Nils Kloweit}: 
    Writing – Original Draft.
\textbf{Stefan Lazarov}: 
    Writing – Original Draft.
\textbf{Michael Lenke}: 
    Writing – Original Draft.
\textbf{Vivien Lohmer}: 
    Writing – Original Draft.
\textbf{Katharina Rohlfing}: 
    Writing – Original Draft.
\textbf{Ingrid Scharlau}: 
    Conceptualization, 
    Writing – Original Draft.
\textbf{Amit Singh}: 
    Writing – Original Draft.
\textbf{Lutz Terfloth}: 
    Conceptualization, 
    Writing – Original Draft.
\textbf{Anna-Lisa Vollmer}: 
    Writing – Original Draft.
\textbf{Yu Wang}: 
    Writing – Original Draft.
\textbf{Annedore Wilmes}: 
    Writing – Original Draft.
\textbf{Britta Wrede}: 
    Writing – Original Draft.

\section*{Acknowledgements}
This research was supported by the German Research Foundation (DFG) in the Collaborative Research Center TRR 318/1 2021 ‘Constructing Explainability’ (438445824).

\bibliographystyle{elsarticle-harv} 
\bibliography{bibliography-csr-article.bib}

\end{document}